\documentclass{article} 
\usepackage{l2cw_preprint,times}

\usepackage{amsmath,amsfonts,bm}

\def\eqref#1{equation~\ref{#1}}

\def\1{\bm{1}}

\DeclareMathAlphabet{\mathsfit}{\encodingdefault}{\sfdefault}{m}{sl}
\SetMathAlphabet{\mathsfit}{bold}{\encodingdefault}{\sfdefault}{bx}{n}

\usepackage{hyperref}
\usepackage{url}
\usepackage{graphicx}
\usepackage{booktabs}
\usepackage{tabularx}
\usepackage{amsmath}
\usepackage{amssymb}
\usepackage{xcolor}

\newcommand{\cost}{\mathrm{cost}}
\newcommand{\len}{\mathrm{len}}

\title{Depot-Closed Multi-Component Construction\\for Neural Vehicle Routing}

\author{
  Shinichiro Hamada \\
  Core Technology, R\&D Division \\
  Panasonic Connect Co., Ltd. \\
  \And
  Hisashi Kashima \\
  Graduate School of Informatics \\
  Kyoto University \\
}

\preprintcopy
\begin{document}

\maketitle

\begin{abstract}
Most neural constructive solvers for the vehicle routing problem (VRP) use route-by-route construction, extending one route until completion before starting the next. This commits route membership early and hinders global coordination across routes. We propose multi-component construction, which maintains many route components simultaneously and merges them in an arbitrary order. This removes the depot-return cue that route-by-route construction obtains from the remaining capacity; to compensate, we introduce an interpretation in which every component is treated as an implicitly depot-closed route. Under this depot-closed interpretation, every intermediate state of standard CVRP construction is a complete feasible solution, and the exact cost reduction of a merge is the Clarke--Wright saving. The neural policy combines this CW-saving signal with the evolving component state to learn \emph{what to connect} and \emph{when to connect}. A policy trained only on CVRP100 outperforms the reported results of representative neural solvers on CVRP100--500 with greedy inference and, reused for ruin-and-reconstruct, performs strongly at all evaluated sizes up to CVRP1000. In a zero-shot Constraint Tightness evaluation with capacities from $C=10$ to $500$, it outperforms the reported neural solvers at every capacity. Controlled analyses show that robustness persists without CW grounding and point to learned route-closing behavior as a plausible contributor to the tight-regime degradation of learned route-by-route solvers.
\end{abstract}

\section{Introduction}
\label{sec:intro}

Neural combinatorial optimization (NCO) solves combinatorial optimization problems with neural networks~\citep{bello2017neural,bengio2021machine,cappart2021combinatorial,wu2026survey}, and routing problems such as the traveling salesman problem (TSP) and the capacitated vehicle routing problem (CVRP) are among its most representative applications. Neural constructive solvers, which generate solutions directly and sequentially, have been particularly successful because they combine fast inference with high solution quality~\citep{vinyals2015pointer,nazari2018reinforcement,kool2019attention,kwon2020pomo,xin2021multi,kim2022symnco,drakulic2023bq,luo2023neural,mnlp,fang2024invit,huang2025reld}.

Most constructive CVRP solvers extend a single active route: they choose the next customer from the current position, return to the depot when the capacity requires it, and then start the next route. In this \emph{route-by-route construction}, the remaining capacity is not only feasibility information but also a cue of how far the current route has grown; because it is synchronized with construction progress, the policy can use it to judge the progress of the route and the timing of the depot return. At the same time, growing one route first means committing its customers one after another before the formation of the other routes can be observed, which limits the global coordination of multiple routes even in standard CVRP.

We therefore revisit route-by-route construction itself. In \emph{multi-component construction}, each customer initially forms a singleton component, and each step connects the endpoints of two different components by one edge. Many route components are maintained simultaneously and grown in an arbitrary order. This makes the commitment order of assignment decisions spanning multiple routes flexible and lets the route partition form while the entire component configuration is observed. This freedom comes at a price: since components grow asynchronously, the unique route progression is lost, and it is no longer obvious how a partial component should relate to the depot as a CVRP route. In our preliminary experiments, a multi-component policy without depot-related grounding formed petal-like route geometry from the coordinates to some extent but closed routes to the depot poorly, suggesting that customer-to-customer geometry alone is insufficient for depot closure.

\begin{figure}[t]
\centering
\includegraphics[width=0.96\linewidth]{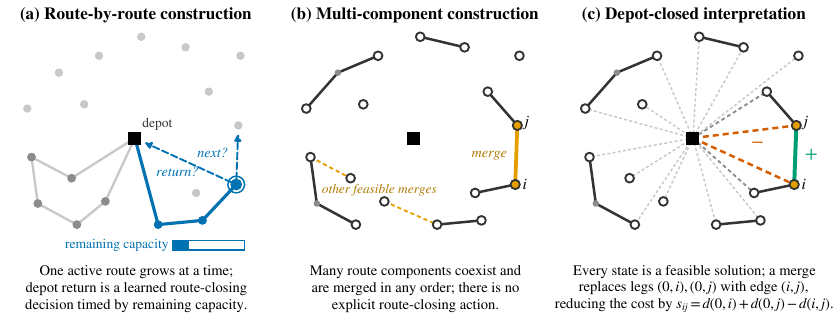}
\caption{Route-by-route versus multi-component construction. (a) Typical learned route-by-route construction grows one active route at a time; remaining capacity provides both feasibility information and a route-progress cue for the learned depot-return decision. (b) Multi-component construction maintains multiple route components simultaneously and selects merges between endpoints of different components; route closure is not an explicit policy action. (c) Under the depot-closed interpretation, every component is treated as an implicitly closed route, so every intermediate state is a feasible CVRP solution and merging endpoints $i$ and $j$ reduces the total cost by exactly the Clarke--Wright saving $s_{ij}=d(0,i)+d(0,j)-d(i,j)$. Here $0$ denotes the depot.}
\label{fig:construction}
\end{figure}

We address this by interpreting each component $P=(v_1,\ldots,v_m)$ not as an open path but as the \emph{implicitly depot-closed route} $0\rightarrow v_1\rightarrow\cdots\rightarrow v_m\rightarrow 0$, which we call the \emph{depot-closed interpretation}. It is closely related to the route-merging construction of the classical Clarke--Wright (CW) savings algorithm~\citep{clarke1964scheduling}, which starts from depot-closed singleton routes and repeatedly merges them. Under this interpretation, every component is a route with a routing cost at that moment, and every intermediate state of standard CVRP construction is a complete feasible solution. Moreover, the exact objective improvement of merging endpoints $i$ and $j$ is the CW saving $s_{ij}=d(0,i)+d(0,j)-d(i,j)$, which our neural policy uses together with the current component state to evaluate candidate merges. Since the order in which edges are committed is not unique even for the same final routes, the policy must also decide when each merge should be executed. We first learn merge connectivity consistent with reference solutions by supervised learning and then improve the entire merge sequence by reinforcement learning under the terminal routing objective, extending learning from \emph{what to connect} to \emph{when to connect} (Figure~\ref{fig:construction}).

We evaluate solution quality on standard CVRP, size generalization from CVRP100 to CVRP200--1000, and zero-shot robustness across capacities on the Constraint Tightness benchmark~\citep{ct}. The final policy outperforms the reported results of representative neural solvers on CVRP100--500 with greedy inference and, reused as a reconstruction operator, performs strongly at all sizes up to CVRP1000, where its greedy inference falls short of LEHD. A single $C=50$ model outperforms the reported neural solvers at every capacity from $C=10$ to $500$; controls in Section~\ref{sec:ct} point to learned route closing as a plausible contributor to the tight-regime failures of learned route-by-route solvers. Our main contributions are as follows.
\begin{enumerate}
\item We propose multi-component construction for NCO, which operates on multiple route components simultaneously in place of route-by-route construction and makes the commitment order of route-assignment decisions spanning multiple routes flexible.
\item To compensate for the depot-return cue lost by moving to multiple components, we introduce the depot-closed interpretation and feed the CW saving, the exact cost reduction of a merge in standard CVRP, into the neural action score as a grounding signal. The depot-closed interpretation keeps the routing cost of every component evaluable, while the CW-saving bias ties neural action selection to the routing objective.
\item We demonstrate strong size and capacity generalization from a single CVRP100 policy, reuse it for reconstruction, and motivate learned route closing as a plausible contributor to tight-regime failures.
\end{enumerate}

\section{Related Work}
\label{sec:related}

\paragraph{Neural constructive solvers for routing.}
Since Pointer Networks, NCO for routing has widely adopted autoregressive construction, which selects the next customer given the current partial solution~\citep{vinyals2015pointer,bello2017neural,nazari2018reinforcement,kool2019attention,kwon2020pomo,xin2021multi}. For CVRP, demands and the remaining capacity enter the decoder context or node features, and solutions are generated route by route. Subsequent work has improved solution quality and generalization through symmetry, state representation, model architecture, and multi-task learning~\citep{kim2022symnco,drakulic2023bq,luo2023neural,mnlp,fang2024invit,huang2025reld,gao2024elg,dgl,zhou2023omni,berto2025routefinder}; multi-task approaches include POMO-MTL~\citep{liu2024mtl} and MVMoE~\citep{zhou2024mvmoe}. Other studies revisit construction itself. JAMPR starts and extends multiple routes simultaneously and selects route--customer pairs as joint actions~\citep{falkner2020learning}, and L2C-Insert inserts an unvisited customer at any valid position of the current partial solution~\citep{l2cinsert}. The TSP decoder of DIFUSCO also greedily joins multiple path fragments~\citep{sun2023difusco}, but its edge scores come from a static heatmap computed before decoding and are not recomputed from an evolving component state. In contrast, our policy uses the current route-component state as its state and selects each merge as a sequential action.

\paragraph{Reconstruction and inference-time refinement.}
Ruin-and-reconstruct reapplies a trained constructive solver to subproblems or partial solutions obtained by ruining part of a solution and is widely used for large-scale generalization and inference-time refinement~\citep{luo2023neural,mnlp,drhg,l2cinsert}. Local rewriting, learned local search, test-time adaptation, and subproblem selection have also been used for refinement in neural routing~\citep{chen2019learning,ma2021learning,hottung2022efficient,li2021learning,ma2023learning}. In our method, construction from singletons and reconstruction from ruined route fragments share the same depot-closed component representation, so the same policy serves both (Section~\ref{sec:rrc}).

\paragraph{Clarke--Wright savings algorithm.}
CW~\citep{clarke1964scheduling} initializes every customer $i$ as a singleton route $0\rightarrow i\rightarrow 0$ and greedily merges routes whose endpoints $i,j$ have a large saving $s_{ij}$ as long as feasibility constraints such as capacity are satisfied; many studies extend the saving criterion or the merging rule~\citep{altinel2005new,enhancedsavings2013}. We call the classical algorithm itself \emph{Classical CW} when it serves as a baseline. Its standard \emph{parallel} variant, used unless otherwise noted, repeatedly performs feasible merges in decreasing order of saving until no merge is possible. Its \emph{sequential} variant extends one active route from both ends in decreasing order of feasible saving and closes it to start the next route only when no feasible extension remains. Although this extension rule differs from the one-ended appending of autoregressive neural solvers, it belongs to the same route-by-route scheme in that only one active route grows at a time; we use it as a control in Section~\ref{sec:ct}. Learning-based enhancements also exist: \citet{legrand_mlcw} use neural edge-selection probabilities to adjust the priority order within the saving list. Rather than learning the ranking rule of the CW heuristic, we incorporate the CW saving as an exact action-local objective signal into a neural merge policy operating on a dynamic multi-component state (Sections~\ref{sec:grounding} and~\ref{sec:disentangle}).

\paragraph{Generalization under size, distribution, and constraint shifts.}
Generalization of neural routing solvers has mainly been studied for size extrapolation and shifts in the node-coordinate distribution~\citep{kim2022symnco,drakulic2023bq,zhou2023omni,luo2023neural,fang2024invit,berto2025routefinder,archgen2025}. The Constraint Tightness study~\citep{ct} shows that merely changing the CVRP capacity substantially changes the solution structure and the required solving strategy, and that models trained with a fixed capacity can overfit to that value; its problem-similarity analysis reports that CVRP100 with $C=10$ is close to OVRP, whereas with $C=500$ it approaches TSP. A capacity shift is thus a structural shift in route count, route geometry, and effective construction strategy, and we use this benchmark in addition to size extrapolation.

\section{Depot-Closed Multi-Component Construction}
\label{sec:method}

\subsection{Multi-component decision process}
\label{sec:mcdp}

Let $0$ denote the depot, $V=\{1,\ldots,N\}$ the customers with locations $\mathbf{x}_i\in\mathbb{R}^2$ and demands $q_i$, and $Q$ the vehicle capacity. Standard CVRP minimizes total length over capacity-feasible depot-to-depot routes, without fixing the vehicle count~\citep{dantzig1959truck,toth2014vehicle}. We represent state $t$ by route components $\mathcal{P}_t=\{P_1,\ldots,P_{K_t}\}$, initially one singleton per customer; each component is a path whose two endpoints alone may participate in the next merge. An action $a_t$ joins endpoints $i,j$ of different components $P_a,P_b$. With $D(P)=\sum_{i\in P}q_i$, the merge is feasible if and only if $D(P_a)+D(P_b)\le Q$, and $\pi_\theta(a\mid\mathcal{P}_t)$ selects among feasible endpoint pairs. Each merge reduces $K_t$ by one. Construction ends when no feasible component pair remains, so route partition and route count emerge jointly from the merge sequence.

This process loses two cues of route-by-route construction: the relation between each partial path and the depot is not explicit, and there is no unique order that determines which feasible merge to commit now. We first address the former.

\subsection{Giving route semantics to each component}
\label{sec:semantics}

Route-by-route construction couples remaining capacity to the progress of one active route; with asynchronous components, this single-route lifecycle disappears. To compensate for this loss, we introduce the depot-closed interpretation: each component $P=(v_1,\ldots,v_m)$ is treated as the implicitly depot-closed route $0\rightarrow v_1\rightarrow\cdots\rightarrow v_m\rightarrow 0$. With the Euclidean distance $d(\cdot,\cdot)$ and the internal path length $\len(P)$, its cost is
\begin{equation}
\cost(P)=d(0,v_1)+\len(P)+d(v_m,0),
\end{equation}
and the cost of a state is $\cost(\mathcal{P}_t)=\sum_{P\in\mathcal{P}_t}\cost(P)$. An important consequence is that, in standard CVRP, every intermediate state is already a complete feasible solution: the initial singleton state contains every customer exactly once, and merging two capacity-feasible components preserves feasibility once each component is closed at the depot. Construction therefore proceeds as a sequence of transformations $\mathcal{P}_t\rightarrow\mathcal{P}_{t+1}$ between feasible solutions, and every partial path has a meaning in terms of the routing objective at every step.

\subsection{Grounding each merge in the routing objective}
\label{sec:grounding}

Merging the endpoints $i,j$ of depot-closed components removes the depot legs $i\rightarrow 0$ and $0\rightarrow j$ and adds the edge $i\rightarrow j$. The objective improvement of a single merge is therefore
\begin{equation}
\cost(\mathcal{P}_t)-\cost(\mathcal{P}_{t+1}) = d(0,i)+d(0,j)-d(i,j) = s_{ij},
\label{eq:saving}
\end{equation}
which is exactly the Clarke--Wright saving. We provide $s_{ij}$ to the policy explicitly because the multi-component representation alone offers no direct point of contact, from the learning perspective, between the depot-closed interpretation and action selection. In supervised learning in particular, the teacher only indicates expert-compatible merges, and the distance change associated with depot closure is never observed; without a depot-aware signal, even a merge rule that imitates customer-to-customer geometry while ignoring depot legs can satisfy the teacher. Since $s_{ij}$ is exactly the cost difference of a merge under the depot-closed interpretation, it is a natural signal for grounding this interpretation in the policy.

Since $\cost(\mathcal{P}_0)=2\sum_i d(0,i)$, the final cost decomposes as $\cost(\mathcal{P}_T)=\cost(\mathcal{P}_0)-\sum_{t=0}^{T-1}s_{a_t}$, where $s_{a_t}$ is the saving of merge $a_t$. The saving thus measures the objective improvement of each merge at that moment, but always selecting the largest current saving is not sufficient: a merge fixes the edges inside the component and changes which endpoint pairs remain selectable, so the policy must also consider which feasible merges will remain in the future. The action score therefore combines two kinds of information. The CW saving exactly gives the immediate improvement of a candidate merge, whereas a learned score conditioned on the current component state supplies the longer-horizon information about future options that the saving lacks, that is, \emph{when to connect}.

\subsection{Global neural reasoning over unordered merges}
\label{sec:arch}

\begin{figure}[t]
\centering
\includegraphics[width=0.86\linewidth]{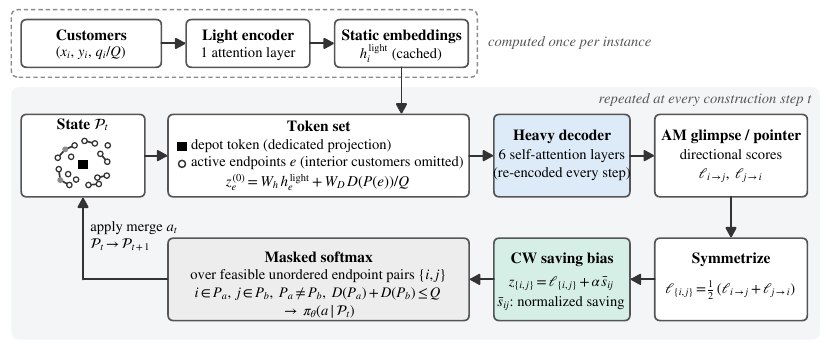}
\caption{Policy architecture. Static customer features are encoded once by a light encoder. At each construction step, the depot token and active endpoints, together with their component loads, are re-encoded by the heavy decoder. An AM-based glimpse/pointer produces directional scores, which are symmetrized into unordered pair scores, augmented by the normalized CW-saving bias, and normalized by a masked softmax over capacity-feasible endpoint pairs. Applying the selected merge produces the next component state. For clarity, the graph-context term used in the full implementation is omitted; the endpoint-token equation is the conceptual form of Section~\ref{sec:arch}.}
\label{fig:architecture}
\end{figure}

To capture longer-horizon effects, the learned score conditions on the full current state. We adapt LEHD's Transformer-based light-encoder/heavy-decoder architecture~\citep{luo2023neural,vaswani2017attention}; its heavy decoder re-encodes the current token set at every construction step (Figure~\ref{fig:architecture}).

\paragraph{Static customer information and dynamic component information.}
The light encoder computes $h_i^{\mathrm{light}}$ once from the static feature $(x_i,y_i,q_i/Q)$. The component load $D(P)/Q$ changes with every merge and is therefore handled by the heavy decoder, which at each step receives only the depot and the active endpoints; internal customers are excluded because they can no longer participate in merges. Conceptually, endpoint $e$ is represented as
\begin{equation}
z_e^{(0)} = W_h h_e^{\mathrm{light}} + W_D\,D(P(e))/Q,
\end{equation}
where $P(e)$ is the component containing $e$. Endpoints share one projection, whereas the depot is a special token with its own projection. The depot is not a merge candidate, but self-attention lets every endpoint refer to both the depot and the other components.

\paragraph{Scoring an unordered endpoint pair.}
With endpoint $i$ as query and $j$ as destination, the Attention Model (AM) glimpse/pointer~\citep{kool2019attention} gives a directional score $\ell_{i\rightarrow j}$. Because merge $\{i,j\}$ is unordered, we symmetrize the two directional scores before the softmax, $\ell_{\{i,j\}}=(\ell_{i\rightarrow j}+\ell_{j\rightarrow i})/2$. Since CW saving is pair-specific, its normalized value enters the pair score as a logit bias rather than a node feature:
\begin{equation}
z_{\{i,j\}}=\ell_{\{i,j\}}+\alpha\,\bar{s}_{ij},
\label{eq:logit}
\end{equation}
where $\bar{s}_{ij}$ is the normalized CW saving and $\alpha$ its strength. A single softmax over all feasible unordered endpoint pairs yields $\pi_\theta(a\mid\mathcal{P}_t)$. The architecture thus separates static customer information in the light encoder, dynamic component state in the heavy decoder, pair compatibility in the pointer, and the objective change in the pair logit.

\subsection{Learning what to connect and when to connect}
\label{sec:learning}

Freeing both what and when to connect from the beginning makes learning difficult because correct connectivity and good merge ordering must be explored simultaneously. We therefore first learn what to connect stably by supervised learning (SL), gradually transfer merge-order freedom to the policy, and finally optimize the entire merge sequence against the terminal routing cost by reinforcement learning (RL) in S3. In parallel, the training states move from intermediate states of ordinary construction (S0--S1) to repair states obtained by partially ruining completed solutions (S2--S3). The curriculum thus has two axes (Table~\ref{tab:curriculum} in Appendix~\ref{app:repro}): decision freedom expands from what to when, and the state distribution expands from construction to reconstruction.

In the SL stages S0--S2, all currently feasible, unselected merge actions corresponding to customer--customer edges in the reference solution are treated as correct. Denoting this set by $A^+(\mathcal{P})$, we impose no unique teacher order and maximize the probability mass assigned to the correct set, $\sum_{a\in A^+(\mathcal{P})}\pi_\theta(a\mid\mathcal{P})$. The stages differ in how the next state is generated. In \emph{shuffle-guided} trajectories (S0), the state is updated by a merge chosen uniformly at random from $A^+(\mathcal{P})$, so the reference-solution edges are committed in a random order independent of the policy. In \emph{policy-guided} trajectories (S1--S2), the action distribution of the policy is restricted to $A^+(\mathcal{P})$; the trajectory remains expert-compatible, but its ordering depends on the current policy. Moving from the random ordering of S0 to the policy-induced ordering of S1 also reduces the mismatch between artificially generated expert-compatible states and the states actually visited by the policy. In all stages, we additionally use \emph{route-subset sampling}: a subset of routes is extracted from a source solution to form a training subproblem, on which the stage-specific initial state is generated. This reduces the training problem size and exposes the policy to different effective problem sizes (Section~\ref{sec:sizeanalysis}).

\subsection{The same policy for construction and reconstruction}
\label{sec:rrc}

To improve solution quality beyond initial construction, we also use the same policy for iterative ruin-and-reconstruct (RRC), which refines an incumbent solution by repeatedly ruining part of it and reconstructing it from the remaining fragments. We use \emph{polar ruin}, inspired by the polar-angle-based destruction strategy used for CVRP in L2C-Insert~\citep{l2cinsert}: the polar angle viewed from the depot defines the neighborhood metric. In each iteration, a customer chosen uniformly at random serves as the ruin center, customers with polar angles close to that of the center are returned to singleton components, and the undestroyed route fragments are kept. The same operator generates the initial states of S2--S3. Regarding each remaining fragment as a depot-closed component, the set of fragments and singletons is a state of exactly the same type as $\mathcal{P}$ in Sections~\ref{sec:mcdp}--\ref{sec:semantics}. Construction and reconstruction therefore differ only in their initial state; the feasible actions, heavy decoder, pair scorer, CW grounding, and policy $\pi_\theta$ are all shared, and no reconstruction-specific solver is needed.

\section{Experiments}
\label{sec:exp}

\subsection{Experimental setup}
\label{sec:setup}

All main models are trained on uniform Euclidean CVRP100 with capacity $C=50$; following the Constraint Tightness benchmark, we write $C$ for the capacity value in the experiments. We follow the protocol of LEHD~\citep{luo2023neural}: training uses 1,000,000 CVRP100 instances solved by Hybrid Genetic Search (HGS)~\citep{vidal2022hybrid}, and the test sets contain 10,000 CVRP100 instances and 128 instances each of CVRP200/500/1000 with capacities 50/80/100/250, with LKH3~\citep{helsgaun2017extension} solutions as references. S0--S3 are successive training stages of the same policy, and S3 is the final model; training settings and runtime results are given in Appendix~\ref{app:repro}. The same policy is applied to all sizes without additional training. Unless otherwise noted, we report the mean per-instance gap $\frac{1}{M}\sum_{m=1}^{M}100\,(L_m-L_m^\ast)/L_m^\ast$ over the $M$ test instances; a negative gap means that the solution is shorter than the reference. Reported values are transcribed from the tables of the cited papers (Source column), and Mean denotes the unweighted average over the four sizes.

\subsection{Standard size generalization}
\label{sec:size}

\begin{table}[t]
\caption{Greedy inference on standard CVRP (mean per-instance gap to LKH3, \%).}
\label{tab:greedy}
\centering
\setlength{\tabcolsep}{2.0pt}
\begin{tabular}{@{}lrrrrrl@{}}
\toprule
Method & CVRP100 & CVRP200 & CVRP500 & CVRP1000 & Mean & Source \\
\midrule
BQ & 2.726 & 2.972 & 3.248 & 5.892 & 3.710 & ReLD Table 3 / BQ original \\
LEHD & 3.648 & 3.312 & 3.178 & \textbf{4.912} & 3.763 & LEHD Table 1; ReLD Table 3 \\
MnLP & 3.509 & 3.206 & 2.928 & 6.195 & 3.960 & MnLP Table 1 \\
INViT & 9.964 & 12.160 & 13.772 & 15.548 & 12.861 & MnLP Table 1 \\
DGL & 6.441 & 8.368 & 9.356 & 16.727 & 10.223 & MnLP Table 1 \\
\midrule
Classical CW & 5.479 & 7.310 & 6.525 & 11.999 & 7.828 & This work \\
S0 & 3.504 & 4.417 & 4.682 & 9.963 & 5.642 & This work \\
S1 & 1.732 & 1.990 & 2.266 & 7.164 & 3.288 & This work \\
S2 & 1.383 & 1.589 & 2.049 & 7.495 & 3.129 & This work \\
S3 & \textbf{1.019} & \textbf{0.828} & \textbf{1.407} & 7.544 & \textbf{2.700} & This work \\
\bottomrule
\end{tabular}
\end{table}

\paragraph{Greedy inference.}
Table~\ref{tab:greedy} applies the policy trained on CVRP100 to CVRP100--1000 with greedy inference; Classical CW uses the parallel variant. As the curriculum proceeds, the mean gap improves monotonically ($5.642\rightarrow3.288\rightarrow3.129\rightarrow2.700$), and the final S3 outperforms the compared reported greedy results on CVRP100, 200, and 500. On CVRP1000, however, greedy inference falls short of LEHD, and the gap even worsens slightly from 7.164\% at S1 to 7.495\%/7.544\% at S2/S3, leaving room for improvement under extreme scale shifts. Classical CW also merges depot-closed routes but selects each merge greedily by saving, and its gap is larger than that of S3 at every size: depot-closed route merging and the CW saving alone cannot explain the performance of S3, and the learned, state-dependent merge selection provides an additional gain. The greedy inference of S3 also runs at a speed comparable to LEHD (Appendix~\ref{app:runtime}).

\begin{table}[t]
\caption{Enhanced inference on standard CVRP (mean per-instance gap to LKH3, \%). Negative values indicate solutions shorter than the LKH3 references.}
\label{tab:enhanced}
\centering
\setlength{\tabcolsep}{2.8pt}
\begin{tabular}{@{}llrrrrrl@{}}
\toprule
Method & Inference & CVRP100 & CVRP200 & CVRP500 & CVRP1000 & Mean & Source \\
\midrule
MDAM & bs50 & 2.211 & 4.304 & 10.498 & 27.814 & 11.207 & ReLD Table 3 \\
POMO & aug$\times$8 & 1.004 & 3.403 & 11.135 & 110.632 & 31.544 & ReLD Table 3 \\
ELG & aug$\times$8 & 1.207 & 2.553 & 5.472 & 10.760 & 4.998 & ReLD Table 3 \\
ReLD & aug$\times$8 & 0.960 & 1.654 & 2.975 & 6.757 & 3.087 & ReLD Table 3 \\
BQ & bs16 & 0.611 & 1.141 & 2.991 & 7.784 & 3.132 & LEHD Table 1 \\
BQ & bs16, later rerun & 1.020 & 0.940 & 1.010 & 2.880 & 1.463 & DRHG Table 3 \\
LEHD & RRC100 & 0.272 & 0.217 & 0.546 & 2.370 & 0.851 & LEHD Table 1 \\
LEHD & RRC1000 & $-$0.112 & $-$0.383 & $-$0.347 & 0.921 & 0.020 & LEHD Table 1 \\
MnLP & RRC1000 & 0.012 & $-$0.255 & $-$0.257 & 1.148 & 0.162 & MnLP Table 1 \\
DRHG & $T{=}1000$ & $-$0.020 & $-$0.160 & 1.310 & 7.660 & 2.198 & DRHG Table 3 \\
\midrule
S0 & RRC1000 & 0.597 & 0.400 & 0.658 & 3.125 & 1.195 & This work \\
S1 & RRC1000 & 0.018 & 0.144 & 0.620 & 1.957 & 0.685 & This work \\
S2 & RRC1000 & $-$0.199 & $-$0.605 & $-$1.102 & $-$0.014 & $-$0.480 & This work \\
S3 & RRC1000 & \textbf{$-$0.216} & \textbf{$-$0.722} & \textbf{$-$1.179} & \textbf{$-$0.272} & \textbf{$-$0.597} & This work \\
\bottomrule
\end{tabular}
\end{table}

\paragraph{Enhanced inference.}
Table~\ref{tab:enhanced} compares the reported settings that use additional inference-time computation (full results in Table~\ref{tab:enhanced_full}). For S0--S3, the same policy serves as the reconstruction operator of RRC with polar ruin and 1,000 reconstruction iterations (RRC1000). Since the other methods use different procedures (augmentation, beam search, or RRC with other ruin operators), the table compares reported enhanced-inference settings rather than a common inference algorithm. RRC resolves the degradation on CVRP1000 that remained under greedy inference: already S2 outperforms the LKH3 references at all four sizes, and S3 outperforms LEHD RRC1000, the best among the compared reconstruction methods, at all four sizes. BQ bs16 appears twice because different sources report different rerun values.

\subsection{Analysis of size generalization}
\label{sec:sizeanalysis}

\begin{table}[t]
\centering
\begin{minipage}[t]{0.56\linewidth}
\centering
\caption{Effect of route-subset sampling (greedy, gap \%).}
\label{tab:subset}
\setlength{\tabcolsep}{3.5pt}
\begin{tabular}{@{}lcrrrrr@{}}
\toprule
Stage & Subset & 100 & 200 & 500 & 1000 & Mean \\
\midrule
S2 & No & 1.217 & 1.833 & 3.615 & 9.952 & 4.154 \\
S2 & Yes & 1.383 & 1.589 & 2.049 & 7.495 & 3.129 \\
S3 & No & 0.889 & 1.159 & 2.746 & 8.057 & 3.213 \\
S3 & Yes & 1.019 & 0.828 & 1.407 & 7.544 & 2.700 \\
\bottomrule
\end{tabular}
\end{minipage}\hfill
\begin{minipage}[t]{0.42\linewidth}
\centering
\caption{CW-grounding bootstrap on CVRP100 (gap \%).}
\label{tab:bootstrap}
\setlength{\tabcolsep}{3.5pt}
\begin{tabular}{@{}llr@{}}
\toprule
Training & Inference & Gap \\
\midrule
no CW grounding & no CW bias & 5.822 \\
no CW grounding & +CW bias & 3.883 \\
CW fine-tuning & +CW bias & \textbf{3.787} \\
\bottomrule
\end{tabular}
\end{minipage}
\end{table}

\paragraph{Effect of route-subset sampling.}
Table~\ref{tab:subset} changes only the presence of route-subset sampling for S2 and S3 while keeping the rest of the training setup fixed. Route-subset sampling sacrifices about 0.1--0.2 pp of the in-distribution gap on CVRP100 but substantially improves extrapolation to CVRP200--1000; on CVRP500, for example, the gap of S3 decreases from 2.746\% to 1.407\%. This suggests that experiencing different effective problem sizes during training, not the architecture alone, is an important factor for size generalization.

\paragraph{Target-size calibration of $\alpha$.}
The saving coefficient $\alpha$ in Eq.~(\ref{eq:logit}) is a scalar hyperparameter rather than a learnable parameter, which allows the strength of the CW-saving signal to be calibrated without retraining the neural policy. In the main results, it is selected on validation data for each target size, since the optimal value differs across sizes. Fixing the CVRP100-selected $\alpha=100$ for all sizes changes the mean gap of S3 only from 2.700\% to 2.725\% and is even slightly better on CVRP500 (Table~\ref{tab:alpha} in Appendix~\ref{app:results}). Target-size-specific calibration is therefore not essential.

\begin{table}[t]
\caption{Zero-shot generalization across constraint tightness on CVRP100 (aggregate gap to HGS, \%). All neural models are trained with $C=50$. Baseline values are taken from the Constraint Tightness benchmark~\citep{ct}; rows below the middle rule are ours. Mean is recomputed from the seven displayed values for all methods.}
\label{tab:ct}
\centering
\setlength{\tabcolsep}{4.0pt}
\begin{tabular}{@{}lrrrrrrrr@{}}
\toprule
Model & $C{=}10$ & $C{=}50$ & $C{=}100$ & $C{=}200$ & $C{=}300$ & $C{=}400$ & $C{=}500$ & Mean \\
\midrule
AM & 45.16 & 7.66 & 12.85 & 21.63 & 26.26 & 25.35 & 27.23 & 23.73 \\
POMO & 20.26 & 3.66 & 11.17 & 22.45 & 29.56 & 30.80 & 34.12 & 21.72 \\
MDAM & 7.47 & 5.38 & 12.47 & 21.13 & 24.31 & 22.55 & 23.91 & 16.75 \\
BQ & 18.02 & 3.25 & 3.48 & 4.53 & 6.53 & 6.54 & 8.16 & 7.22 \\
LEHD & 36.56 & 4.22 & 4.87 & 4.18 & 5.78 & 4.92 & 6.82 & 9.62 \\
ELG & 10.44 & 5.24 & 10.91 & 18.70 & 22.81 & 21.84 & 24.05 & 16.28 \\
INViT & 17.44 & 7.85 & 11.12 & 14.36 & 15.86 & 12.61 & 13.71 & 13.28 \\
POMO-MTL & 13.62 & 4.50 & 8.20 & 10.40 & 14.12 & 13.44 & 15.57 & 11.41 \\
MVMoE & 10.72 & 5.06 & 10.52 & 17.63 & 21.80 & 21.86 & 16.00 & 14.80 \\
\midrule
Classical CW (parallel) & 4.31 & 5.95 & 9.92 & 11.31 & 10.78 & 9.03 & 8.19 & 8.50 \\
Classical CW (sequential) & 5.59 & 16.44 & 23.64 & 24.02 & 21.31 & 18.57 & 14.68 & 17.75 \\
No-CW base & 5.91 & 6.28 & 7.94 & 8.04 & 8.11 & 7.45 & 7.56 & 7.33 \\
S0 & 4.81 & 4.00 & 5.80 & 6.01 & 5.90 & 4.75 & 4.86 & 5.16 \\
S1 & \textbf{4.17} & 2.23 & 3.22 & 3.98 & 4.62 & 4.13 & 4.60 & 3.85 \\
S2 & 5.49 & 1.87 & 2.75 & 3.17 & 3.78 & 3.50 & 4.05 & 3.52 \\
S3 & 5.42 & \textbf{1.51} & \textbf{2.13} & \textbf{2.55} & \textbf{3.04} & \textbf{2.92} & \textbf{3.40} & \textbf{3.00} \\
\bottomrule
\end{tabular}
\end{table}

\subsection{Disentangling neural reasoning and Clarke--Wright grounding}
\label{sec:disentangle}

To separate the learned score from the analytic CW saving, besides the gap between Classical CW and S3 (Table~\ref{tab:greedy}), we decompose the CW-grounding bootstrap that forms S0 (Table~\ref{tab:bootstrap}): a base policy is trained by shuffle-guided SL with $\alpha=0$, $\alpha$ is selected on validation data, and the policy is then fine-tuned on the CW-grounded logits. This diagnostic comparison, which does not use the final S0 checkpoint of Table~\ref{tab:greedy}, separates the effect of adding the CW signal post hoc from that of adapting the policy to the resulting action landscape. Adding the signal post hoc improves the gap from 5.822\% to 3.883\%, so making the objective change of each candidate merge explicit contributes substantially on its own; fine-tuning further improves it to 3.787\%, showing that the learned score also adapts to the CW-grounded action space. Together with the large gap between Classical CW and S3, both the analytic local objective signal and the learned state-dependent merge scheduling contribute to solution quality. Section~\ref{sec:ct} shows, however, that basic constraint robustness appears even before CW grounding is introduced.

\subsection{Generalization across constraint tightness}
\label{sec:ct}

Following the Constraint Tightness benchmark~\citep{ct}, we fix the problem size to CVRP100 and evaluate $C\in\{10,50,100,200,300,400,500\}$ zero-shot: all neural models are trained with $C=50$, and $\alpha$ is not re-tuned per capacity (each model keeps the value selected at $C=50$). Following the original protocol, we report the aggregate gap $100\,(\sum_m L_m-\sum_m L_m^\ast)/\sum_m L_m^\ast$ with HGS as the reference.

S3 outperforms the reported neural solvers at every capacity, with a mean gap of 3.00\% versus 7.22\% for BQ and 9.62\% for LEHD (Table~\ref{tab:ct}). The difference is largest in the tight regime: from $C=50$ to $C=10$, LEHD degrades from 4.22\% to 36.56\% and BQ from 3.25\% to 18.02\%, whereas S3 increases from 1.51\% to 5.42\%. The stage-wise improvement is not monotone at every capacity; the best value at $C=10$ is 4.17\% (S1).

To diagnose the tight-regime failure mode---not to establish robustness from one capacity alone---we focus on $C=10$, where the separation is sharpest; Table~\ref{tab:ct} provides the full sweep. At $C=10$, every reported learned route-by-route solver has a gap of at least 7.47\%, whereas both non-learned CW controls and all our multi-component policies are at most 5.91\%. No single attribute explains this split. CW saving does not: parallel CW reaches 4.31\%, while the No-CW base uses no saving and reaches 5.91\%. Route-by-route construction does not either: sequential CW grows one route at a time yet reaches 5.59\%. Nor does learning alone, since the No-CW base and S0--S3 are learned. The sharpest contrast is LEHD versus the No-CW base: both use the light-encoder/heavy-decoder design, supervised training on HGS solutions, and no saving signal, yet they reach 36.56\% and 5.91\% at $C=10$. Among the evaluated methods, the combination of learning and route-by-route construction coincides exactly with the poorly performing group.

This conjunction has a natural mechanistic interpretation: learned route closing. In learned route-by-route solvers, the depot-return decision is learned under the state progression of the training capacity, where remaining capacity can also cue route progress; a large capacity shift can therefore create a mismatch at test time. Our policy has no route-closing action, while sequential CW closes a route deterministically when no feasible extension remains. We therefore hypothesize that learned route closing contributes to the tight-regime degradation. The full sweep is qualitatively consistent: several learned route-by-route baselines deteriorate again for $C>50$ as routes lengthen, although BQ and LEHD remain comparatively stable. The controls do not isolate this cause: the same poorly performing group is also the set of route-by-route methods without an analytic saving signal, and other capacity-dependent learned behaviors may contribute.

Separately, the CW controls show that construction scheme strongly affects solution quality. Under the same saving rule, sequential CW is 6.49--13.72 percentage points worse than parallel CW for $C=50$--$500$ (mean 17.75\% vs.\ 8.50\%), but only 1.28 points worse at $C=10$ (5.59\% vs.\ 4.31\%). Thus route-by-route construction alone does not explain the severe tight-regime degradation, while construction scheme still has a large effect when routes become long.

\section{Conclusion}
\label{sec:conclusion}

We proposed multi-component construction, which changes the basic unit of neural route construction from a single active route to multiple simultaneous route components. Interpreting every component as a depot-closed route makes every intermediate state of standard CVRP a complete feasible solution and makes the objective improvement of a merge equal to the Clarke--Wright saving; the neural policy combines this signal with the evolving component state to learn which components to connect and when. A single CVRP100 policy performed strongly under size and capacity shifts. Controls point to learned route closing as a plausible contributor to tight-regime failures, while separate CW controls show that construction scheme strongly affects solution quality.

\paragraph{Limitations.} Greedy inference on CVRP1000 falls short of LEHD. Edges once fixed inside a component cannot be changed directly during ordinary construction, so major corrections require re-partitioning by ruin-and-reconstruct. We focused on additive total-distance CVRP with a variable number of routes; prescribed route counts and non-additive route-level objectives such as min-max routing are left for future work. Nevertheless, our results suggest that multi-component construction is a promising alternative paradigm for neural route construction.

\subsection*{AI use statement}
Generative AI tools, primarily ChatGPT and Claude, were used as research assistants throughout this work. Their use included technical discussion and brainstorming during the development of the method, literature search and reference verification, assistance with software implementation and debugging, discussion of experimental design and analysis of experimental results, and assistance in drafting and preparing the manuscript.

In particular, the tools were used to discuss alternative modeling and algorithmic choices, review and generate portions of experimental code and command-line procedures, help diagnose implementation and experimental issues, and assist in organizing and interpreting experimental results. They were also used to improve the presentation of the work, including structuring the manuscript, assisting with English translation and editing, and assisting with \LaTeX{} preparation.

AI-generated suggestions and outputs were reviewed and validated by the authors as appropriate. Final decisions regarding the research questions, proposed method, experimental protocol, interpretation of results, and conclusions were made by the authors. The authors take full responsibility for the correctness and final content of this work, including content produced with the assistance of generative AI.

\subsection*{Reproducibility statement}
We provide the details required to reproduce the proposed method and experiments throughout the paper and appendix. The multi-component decision process, depot-closed representation, policy architecture and action scoring, training curriculum, and reconstruction procedure are specified in Sections~\ref{sec:mcdp}--\ref{sec:rrc}. The datasets, evaluation protocol, reference solutions, evaluation metrics, and main experimental settings are described in Section~\ref{sec:setup} and the corresponding experimental subsections (Sections~\ref{sec:size}--\ref{sec:ct}). Detailed model and training configurations, including architecture hyperparameters, stage-wise optimization settings, random seeds, inference settings, and the runtime measurement protocol, are provided in Appendix~\ref{app:repro}.

\bibliography{references}

@inproceedings{bello2017neural,
  title     = {Neural Combinatorial Optimization with Reinforcement Learning},
  author    = {Bello, Irwan and Pham, Hieu and Le, Quoc V. and Norouzi, Mohammad and Bengio, Samy},
  booktitle = {International Conference on Learning Representations (Workshop Track)},
  year      = {2017}
}

@article{bengio2021machine,
  title   = {Machine Learning for Combinatorial Optimization: A Methodological Tour d'Horizon},
  author  = {Bengio, Yoshua and Lodi, Andrea and Prouvost, Antoine},
  journal = {European Journal of Operational Research},
  volume  = {290},
  number  = {2},
  pages   = {405--421},
  year    = {2021}
}

@inproceedings{cappart2021combinatorial,
  title     = {Combinatorial Optimization and Reasoning with Graph Neural Networks},
  author    = {Cappart, Quentin and Ch{\'e}telat, Didier and Khalil, Elias B. and Lodi, Andrea and Morris, Christopher and Veli{\v{c}}kovi{\'c}, Petar},
  booktitle = {Proceedings of the International Joint Conference on Artificial Intelligence (IJCAI), Survey Track},
  year      = {2021}
}

@article{wu2026survey,
  title   = {Neural Combinatorial Optimization Algorithms for Solving Vehicle Routing Problems: A Comprehensive Survey With Perspectives},
  author  = {Wu, Xuan and Wen, Lijie and Xiao, Yubin and Wu, Chunguo and Wu, Yuesong and Yu, Chaoyu and Maskell, Douglas L. and Zhou, You and Wang, Di},
  journal = {IEEE Transactions on Neural Networks and Learning Systems},
  year    = {2026},
  doi     = {10.1109/TNNLS.2026.3713193}
}

@inproceedings{vinyals2015pointer,
  title     = {Pointer Networks},
  author    = {Vinyals, Oriol and Fortunato, Meire and Jaitly, Navdeep},
  booktitle = {Advances in Neural Information Processing Systems},
  year      = {2015}
}

@inproceedings{nazari2018reinforcement,
  title     = {Reinforcement Learning for Solving the Vehicle Routing Problem},
  author    = {Nazari, Mohammadreza and Oroojlooy, Afshin and Snyder, Lawrence and Tak{\'a}{\v{c}}, Martin},
  booktitle = {Advances in Neural Information Processing Systems},
  year      = {2018}
}

@inproceedings{kool2019attention,
  title     = {Attention, Learn to Solve Routing Problems!},
  author    = {Kool, Wouter and van Hoof, Herke and Welling, Max},
  booktitle = {International Conference on Learning Representations},
  year      = {2019}
}

@inproceedings{kwon2020pomo,
  title     = {{POMO}: Policy Optimization with Multiple Optima for Reinforcement Learning},
  author    = {Kwon, Yeong-Dae and Choo, Jinho and Kim, Byoungjip and Yoon, Iljoo and Gwon, Youngjune and Min, Seungjai},
  booktitle = {Advances in Neural Information Processing Systems},
  year      = {2020}
}

@inproceedings{xin2021multi,
  title     = {Multi-Decoder Attention Model with Embedding Glimpse for Solving Vehicle Routing Problems},
  author    = {Xin, Liang and Song, Wen and Cao, Zhiguang and Zhang, Jie},
  booktitle = {Proceedings of the AAAI Conference on Artificial Intelligence},
  year      = {2021}
}

@inproceedings{kim2022symnco,
  title     = {{Sym-NCO}: Leveraging Symmetricity for Neural Combinatorial Optimization},
  author    = {Kim, Minsu and Park, Junyoung and Park, Jinkyoo},
  booktitle = {Advances in Neural Information Processing Systems},
  year      = {2022}
}

@inproceedings{drakulic2023bq,
  title     = {{BQ-NCO}: Bisimulation Quotienting for Efficient Neural Combinatorial Optimization},
  author    = {Drakulic, Darko and Michel, Sofia and Mai, Florian and Sors, Arnaud and Andreoli, Jean-Marc},
  booktitle = {Advances in Neural Information Processing Systems},
  year      = {2023}
}

@inproceedings{luo2023neural,
  title     = {Neural Combinatorial Optimization with Heavy Decoder: Toward Large Scale Generalization},
  author    = {Luo, Fu and Lin, Xi and Liu, Fei and Zhang, Qingfu and Wang, Zhenkun},
  booktitle = {Advances in Neural Information Processing Systems},
  year      = {2023}
}

@inproceedings{mnlp,
  title     = {Learning with Foresight: Enhancing Neural Routing Policy via Multi-Node Lookahead Prediction},
  author    = {Jiang, Xia and Wu, Yaoxin and Ong, Yew-Soon and Zhang, Yingqian},
  booktitle = {Proceedings of the Thirty-Fifth International Joint Conference on Artificial Intelligence},
  pages     = {6166--6174},
  year      = {2026},
  doi       = {10.24963/ijcai.2026/686}
}

@inproceedings{fang2024invit,
  title     = {{INViT}: A Generalizable Routing Problem Solver with Invariant Nested View Transformer},
  author    = {Fang, Han and Song, Zhihao and Weng, Paul and Ban, Yutong},
  booktitle = {International Conference on Machine Learning},
  year      = {2024}
}

@inproceedings{huang2025reld,
  title     = {Rethinking Light Decoder-Based Solvers for Vehicle Routing Problems},
  author    = {Huang, Ziwei and Zhou, Jianan and Cao, Zhiguang and Xu, Yixin},
  booktitle = {International Conference on Learning Representations},
  year      = {2025}
}

@inproceedings{gao2024elg,
  title     = {Towards Generalizable Neural Solvers for Vehicle Routing Problems via Ensemble with Transferrable Local Policy},
  author    = {Gao, Chengrui and Shang, Haopu and Xue, Ke and Li, Dong and Qian, Chao},
  booktitle = {Proceedings of the Thirty-Third International Joint Conference on Artificial Intelligence},
  pages     = {6914--6922},
  year      = {2024},
  doi       = {10.24963/ijcai.2024/764}
}

@inproceedings{dgl,
  title     = {{DGL}: Dynamic Global-Local Information Aggregation for Scalable {VRP} Generalization with Self-Improvement Learning},
  author    = {Xiao, Yubin and Wu, Yuesong and Cao, Rui and Wang, Di and Cao, Zhiguang and Wu, Xuan and Zhao, Peng and Li, Yuanshu and Zhou, You and Jiang, Yuan},
  booktitle = {Proceedings of the Thirty-Fourth International Joint Conference on Artificial Intelligence},
  pages     = {8669--8677},
  year      = {2025},
  doi       = {10.24963/ijcai.2025/964}
}

@inproceedings{zhou2023omni,
  title     = {Towards Omni-generalizable Neural Methods for Vehicle Routing Problems},
  author    = {Zhou, Jianan and Wu, Yaoxin and Song, Wen and Cao, Zhiguang and Zhang, Jie},
  booktitle = {International Conference on Machine Learning},
  year      = {2023}
}

@article{berto2025routefinder,
  title   = {{RouteFinder}: Towards Foundation Models for Vehicle Routing Problems},
  author  = {Berto, Federico and Hua, Chuanbo and Zepeda, Nayeli Gast and Hottung, Andr{\'e} and Wouda, Niels A. and Lan, Leon and Park, Junyoung and Tierney, Kevin and Park, Jinkyoo},
  journal = {Transactions on Machine Learning Research},
  volume  = {2025},
  number  = {9},
  pages   = {1--19},
  year    = {2025},
  issn    = {2835-8856}
}

@inproceedings{liu2024mtl,
  title     = {Multi-Task Learning for Routing Problem with Cross-Problem Zero-Shot Generalization},
  author    = {Liu, Fei and Lin, Xi and Wang, Zhenkun and Zhang, Qingfu and Tong, Xialiang and Yuan, Mingxuan},
  booktitle = {Proceedings of the 30th ACM SIGKDD Conference on Knowledge Discovery and Data Mining},
  pages     = {1898--1908},
  year      = {2024},
  doi       = {10.1145/3637528.3672040}
}

@inproceedings{zhou2024mvmoe,
  title     = {{MVMoE}: Multi-Task Vehicle Routing Solver with Mixture-of-Experts},
  author    = {Zhou, Jianan and Cao, Zhiguang and Wu, Yaoxin and Song, Wen and Ma, Yining and Zhang, Jie and Chi, Xu},
  booktitle = {Proceedings of the 41st International Conference on Machine Learning},
  series    = {Proceedings of Machine Learning Research},
  volume    = {235},
  pages     = {61804--61824},
  year      = {2024}
}

@article{archgen2025,
  title   = {Improving Generalization of Neural Vehicle Routing Problem Solvers Through the Lens of Model Architecture},
  author  = {Xiao, Yubin and Wang, Di and Wu, Xuan and Wu, Yuesong and Li, Boyang and Du, Wei and Wang, Liupu and Zhou, You},
  journal = {Neural Networks},
  volume  = {187},
  pages   = {107380},
  year    = {2025},
  doi     = {10.1016/j.neunet.2025.107380}
}

@inproceedings{ct,
  title     = {Rethinking Neural Combinatorial Optimization for Vehicle Routing Problems with Different Constraint Tightness Degrees},
  author    = {Luo, Fu and Wu, Yaoxin and Zheng, Zhi and Wang, Zhenkun},
  booktitle = {Advances in Neural Information Processing Systems},
  volume    = {38},
  year      = {2025},
  doi       = {10.52202/085713-4240}
}

@article{falkner2020learning,
  title   = {Learning to Solve Vehicle Routing Problems with Time Windows through Joint Attention},
  author  = {Falkner, Jonas K. and Schmidt-Thieme, Lars},
  journal = {arXiv preprint arXiv:2006.09100},
  year    = {2020}
}

@inproceedings{l2cinsert,
  title     = {Learning to Insert for Constructive Neural Vehicle Routing Solver},
  author    = {Luo, Fu and Lin, Xi and Zhong, Mengyuan and Liu, Fei and Wang, Zhenkun and Sun, Jianyong and Zhang, Qingfu},
  booktitle = {Advances in Neural Information Processing Systems},
  volume    = {38},
  year      = {2025},
  doi       = {10.52202/085713-2918}
}

@inproceedings{sun2023difusco,
  title     = {{DIFUSCO}: Graph-based Diffusion Solvers for Combinatorial Optimization},
  author    = {Sun, Zhiqing and Yang, Yiming},
  booktitle = {Advances in Neural Information Processing Systems},
  year      = {2023}
}

@article{drhg,
  title   = {Destroy and Repair Using Hyper-Graphs for Routing},
  author  = {Li, Ke and Liu, Fei and Wang, Zhenkun and Zhang, Qingfu},
  journal = {Proceedings of the AAAI Conference on Artificial Intelligence},
  volume  = {39},
  number  = {17},
  pages   = {18341--18349},
  year    = {2025},
  doi     = {10.1609/aaai.v39i17.34018}
}

@inproceedings{chen2019learning,
  title     = {Learning to Perform Local Rewriting for Combinatorial Optimization},
  author    = {Chen, Xinyun and Tian, Yuandong},
  booktitle = {Advances in Neural Information Processing Systems},
  year      = {2019}
}

@inproceedings{ma2021learning,
  title     = {Learning to Iteratively Solve Routing Problems with Dual-Aspect Collaborative Transformer},
  author    = {Ma, Yining and Li, Jingwen and Cao, Zhiguang and Song, Wen and Zhang, Le and Chen, Zhenghua and Tang, Jing},
  booktitle = {Advances in Neural Information Processing Systems},
  year      = {2021}
}

@inproceedings{hottung2022efficient,
  title     = {Efficient Active Search for Combinatorial Optimization Problems},
  author    = {Hottung, Andr{\'e} and Kwon, Yeong-Dae and Tierney, Kevin},
  booktitle = {International Conference on Learning Representations},
  year      = {2022}
}

@inproceedings{li2021learning,
  title     = {Learning to Delegate for Large-scale Vehicle Routing},
  author    = {Li, Sirui and Yan, Zhongxia and Wu, Cathy},
  booktitle = {Advances in Neural Information Processing Systems},
  year      = {2021}
}

@inproceedings{ma2023learning,
  title     = {Learning to Search Feasible and Infeasible Regions of Routing Problems with Flexible Neural k-Opt},
  author    = {Ma, Yining and Cao, Zhiguang and Chee, Yeow Meng},
  booktitle = {Advances in Neural Information Processing Systems},
  year      = {2023}
}

@article{clarke1964scheduling,
  title   = {Scheduling of Vehicles from a Central Depot to a Number of Delivery Points},
  author  = {Clarke, G. and Wright, J. W.},
  journal = {Operations Research},
  volume  = {12},
  number  = {4},
  pages   = {568--581},
  year    = {1964}
}

@article{altinel2005new,
  title   = {A New Enhancement of the {Clarke} and {Wright} Savings Heuristic for the Capacitated Vehicle Routing Problem},
  author  = {Alt{\i}nel, {\.I}. Kuban and {\"O}ncan, Temel},
  journal = {Journal of the Operational Research Society},
  volume  = {56},
  number  = {8},
  pages   = {954--961},
  year    = {2005}
}

@article{enhancedsavings2013,
  title   = {Enhanced Savings Calculation and Its Applications for Solving Capacitated Vehicle Routing Problem},
  author  = {Stanojevi{\'c}, Milan and Stanojevi{\'c}, Bogdana and Vujo{\v{s}}evi{\'c}, Mirko},
  journal = {Applied Mathematics and Computation},
  volume  = {219},
  number  = {20},
  pages   = {10302--10312},
  year    = {2013},
  doi     = {10.1016/j.amc.2013.04.002}
}

@inproceedings{legrand_mlcw,
  title     = {Using Machine Learning to Enhance {Clarke} and {Wright} Heuristic},
  author    = {Legrand, Cl{\'e}ment and Accorsi, Luca and Cattaruzza, Diego and Jourdan, Laetitia and Kessaci, Marie-El{\'e}onore and Vigo, Daniele},
  booktitle = {ROADEF 2021},
  year      = {2021},
  note      = {Conference communication}
}

@article{dantzig1959truck,
  title   = {The Truck Dispatching Problem},
  author  = {Dantzig, G. B. and Ramser, J. H.},
  journal = {Management Science},
  volume  = {6},
  number  = {1},
  pages   = {80--91},
  year    = {1959}
}

@book{toth2014vehicle,
  title     = {Vehicle Routing: Problems, Methods, and Applications},
  editor    = {Toth, Paolo and Vigo, Daniele},
  edition   = {2nd},
  publisher = {SIAM},
  year      = {2014}
}

@article{vidal2022hybrid,
  title   = {Hybrid Genetic Search for the {CVRP}: Open-Source Implementation and {SWAP*} Neighborhood},
  author  = {Vidal, Thibaut},
  journal = {Computers \& Operations Research},
  volume  = {140},
  pages   = {105643},
  year    = {2022}
}

@techreport{helsgaun2017extension,
  title       = {An Extension of the {Lin-Kernighan-Helsgaun} {TSP} Solver for Constrained Traveling Salesman and Vehicle Routing Problems},
  author      = {Helsgaun, Keld},
  institution = {Roskilde University},
  year        = {2017}
}

@inproceedings{vaswani2017attention,
  title     = {Attention Is All You Need},
  author    = {Vaswani, Ashish and Shazeer, Noam and Parmar, Niki and Uszkoreit, Jakob and Jones, Llion and Gomez, Aidan N. and Kaiser, {\L}ukasz and Polosukhin, Illia},
  booktitle = {Advances in Neural Information Processing Systems},
  year      = {2017}
}
\bibliographystyle{iclr2027_conference}

\appendix
\section{Reproducibility and runtime}
\label{app:repro}

\subsection{Model and training configuration}

\begin{table}[h]
\caption{Training curriculum. All stages train the same policy, starting from the checkpoint of the preceding stage.}
\label{tab:curriculum}
\centering
\setlength{\tabcolsep}{3.5pt}
\begin{tabularx}{\linewidth}{@{}lllX@{}}
\toprule
Stage & Trajectory; initial state & Objective & Role \\
\midrule
S0 & shuffle-guided; construction prefix & set-valued SL & Shuffling suppresses dependence on a particular merge ordering; focuses on valid merge selection (what to connect). \\
S1 & policy-guided; construction prefix & set-valued SL & Learns what to connect on expert-compatible, policy-generated construction trajectories. \\
S2 & policy-guided; ruined solution & set-valued SL & Learns what to connect on expert-compatible, policy-generated repair trajectories. \\
S3 & free policy rollout; ruined solution & terminal route-cost RL & Removes the teacher constraint and jointly learns what and when to connect. \\
\bottomrule
\end{tabularx}
\end{table}

All stages use the same network architecture. The static customer feature $(x,y,q/Q)$ is mapped to an embedding dimension of 128 and encoded once by a one-layer light encoder. The component state, which changes during construction, is re-encoded by a six-layer heavy decoder. Multi-head attention uses 8 heads with a per-head Q/K/V dimension of 16 and a feed-forward hidden dimension of 512, and we use the pointer mechanism and graph context of the Attention Model (AM)~\citep{kool2019attention}. The model has 1.56M parameters (Table~\ref{tab:arch}).

\begin{table}[h]
\caption{Model configuration.}
\label{tab:arch}
\centering
\begin{tabular}{@{}ll@{}}
\toprule
Item & Setting \\
\midrule
Training problem & CVRP100, $C=50$ \\
Training / validation data & 1,000,000 HGS-solved instances / 1,000 instances \\
Embedding dimension & 128 \\
Light encoder & 1 attention layer \\
Heavy decoder & 6 attention layers \\
Attention & 8 heads, Q/K/V dimension 16 per head \\
Feed-forward hidden dimension & 512 \\
Pointer / context & AM pointer mechanism; graph context \\
Parameters & 1,562,752 \\
Saving normalization & instance-wise maximum \\
Random seed & 22 \\
\bottomrule
\end{tabular}
\end{table}

Training follows the S0--S3 curriculum, and each stage inherits the validation-selected checkpoint of the preceding stage. S0--S2 use supervised learning with Adam, and S3 uses REINFORCE with a rollout baseline. S0--S2 use centered-logit $L_2$ regularization ($\lambda=3\times10^{-4}$). The rollout baseline of S3 is initialized from the S2 policy and updated when the candidate policy improves significantly on a 1,000-instance challenge set. Table~\ref{tab:stages} lists the stage-specific settings.

\begin{table}[h]
\caption{Stage-specific training settings. ``constr.\ prefix'' denotes a (shuffled or policy-generated) construction prefix; the S3 batch of 512 is processed in micro-batches of 256.}
\label{tab:stages}
\centering
\setlength{\tabcolsep}{3.5pt}
\begin{tabular}{@{}lllll@{}}
\toprule
Setting & S0 & S1 & S2 & S3 \\
\midrule
Learning objective & set-valued SL & set-valued SL & set-valued SL & REINFORCE \\
State generation & shuffle-guided & policy-guided & policy-guided & free rollout \\
Initial state & constr.\ prefix & constr.\ prefix & polar ruin & polar ruin \\
Route-subset sampling & Yes & Yes & Yes & Yes \\
Optimizer & Adam & Adam & Adam & Adam \\
Learning rate & $10^{-4}$ & $10^{-4}$ & $10^{-4}$ & $10^{-5}$ \\
LR decay factor & 0.9 & 0.9 & 0.9 & 1.0 \\
Training batch size & 512 & 512 & 512 & 512 (micro 256) \\
Gradient clipping & None & None & 5 & 1 \\
Training $\alpha$ & 72 & 85 & 95 & 100 \\
Centered-logit $L_2$ & $3\times10^{-4}$ & $3\times10^{-4}$ & $3\times10^{-4}$ & -- \\
Ruin metric / range & -- & -- & polar / 100 & polar / 100 \\
\bottomrule
\end{tabular}
\end{table}

For polar ruin, a customer is selected uniformly at random as the ruin center. The destroy size is sampled from $\{4,\ldots,\min(100,n)\}$, and the customers with the closest polar angles to the center are removed. At test time, greedy inference performs a single construction pass from singleton components. RRC1000 uses the same policy with 1,000 reconstruction iterations; for a problem of size $n$, its destroy-size range is $\{4,\ldots,n\}$, and the reported runs use seed 4567.

For standard size generalization, the inference-time saving coefficient is selected on validation data separately for each training stage and target size. For the final S3 model used in the headline results, the selected values are $\alpha=100,100,105,200$ for CVRP100/200/500/1000, respectively. The fixed-$\alpha$ ablation uses the CVRP100-selected value $\alpha=100$ for all four sizes. In the Constraint Tightness evaluation, each model uses a single $\alpha$ selected at the training capacity $C=50$ and keeps it fixed across all target capacities; no target-capacity-specific calibration is performed.

\subsection{Inference runtime}
\label{app:runtime}

Runtime is measured on a single NVIDIA RTX PRO 6000 Blackwell Max-Q Workstation Edition GPU. For direct same-environment comparison, we rerun LEHD greedy inference on the same test instances as S3. We report the wall-clock time measured by each evaluation driver over its inference region, after model and dataset loading and before result serialization, divided by the number of evaluated instances. For both S3 and the LEHD greedy rerun, the evaluation batch sizes are 1000 for CVRP100 and 128 for CVRP200/500/1000. All same-environment measurements use the same machine and software environment.

\begin{table}[h]
\caption{Inference runtime on the same RTX PRO 6000 system (wall-clock seconds per instance).}
\label{tab:runtime}
\centering
\setlength{\tabcolsep}{5.0pt}
\begin{tabular}{@{}llrrrr@{}}
\toprule
Method & Inference & CVRP100 & CVRP200 & CVRP500 & CVRP1000 \\
\midrule
LEHD (rerun) & greedy  & 0.0012 & 0.0116 & 0.0628 & 0.3845 \\
S3           & greedy  & 0.0024 & 0.0156 & 0.0611 & 0.3115 \\
S3           & RRC1000 & 1.176  & 8.354  & 37.783 & 204.096 \\
\bottomrule
\end{tabular}
\end{table}

Greedy inference is in the same runtime regime as LEHD: S3 is slower on CVRP100 and CVRP200, nearly identical on CVRP500, and faster on CVRP1000. RRC1000 is substantially more expensive because it performs 1,000 destroy--repair iterations; Table~\ref{tab:runtime} therefore reports its absolute inference cost rather than treating the iteration count as a compute-matched budget.

We also attempted a same-environment rerun of LEHD RRC1000. Unlike the greedy rerun, which reproduces the published LEHD greedy quality closely (3.636\% versus 3.648\% on CVRP100), the RRC rerun did not reproduce the published RRC solution quality in our environment. We therefore do not use its timing as a valid LEHD RRC runtime measurement. For context only, the original LEHD study reports total CVRP RRC1000 runtimes of 2.8~h, 11.3~min, 2~h, and 8~h over the full CVRP100/200/500/1000 test sets (10,000 and 128 instances), corresponding to roughly 1.0, 5.3, 56, and 225~s per instance, respectively, on a single NVIDIA GeForce RTX 3090 GPU~\citep{luo2023neural}. These published values were obtained under different hardware and implementation conditions and are not used for strict runtime ranking against S3.

\begin{figure}[h!]
\centering
\includegraphics[width=0.98\linewidth]{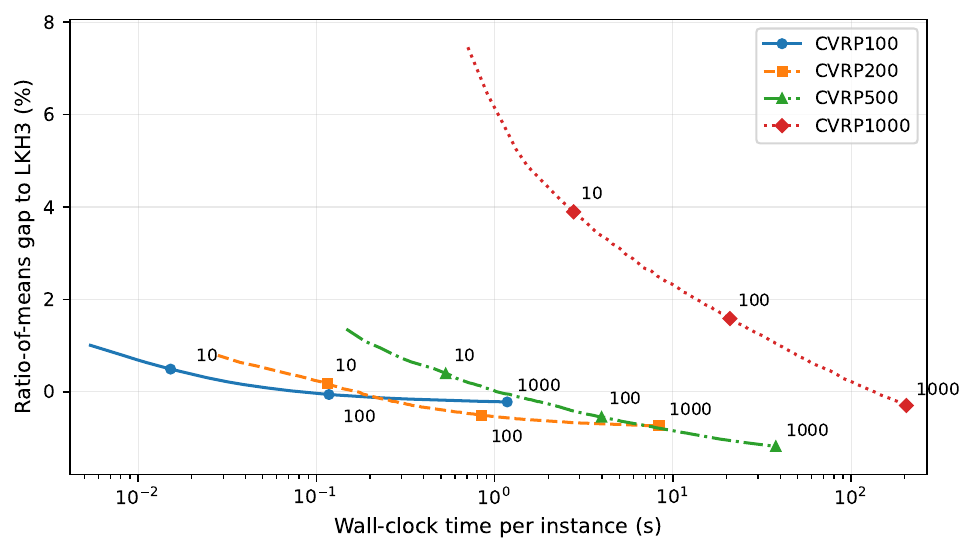}
\caption{Anytime quality--runtime trade-off of S3 with polar RRC. The x-axis shows wall-clock time per instance on a logarithmic scale; markers are labeled by reconstruction iteration (10, 100, and 1,000). The y-axis is the ratio-of-means gap to the LKH3 references computed from the saved incumbent-cost trajectory. Lower is better.}
\label{fig:rrc_anytime}
\end{figure}

Figure~\ref{fig:rrc_anytime} characterizes the compute--quality trade-off within S3 itself using the saved trajectory of the same polar-RRC runs. The x-axis is wall-clock time per instance, reconstructed from the measured cumulative RRC time at each iteration plus the fixed per-run overhead so that the RRC1000 endpoint coincides with Table~\ref{tab:runtime}. Because the saved trajectory contains the mean incumbent cost rather than the per-instance gaps, the y-axis uses the ratio-of-means gap to the LKH3 references rather than the mean per-instance gap used in Tables~\ref{tab:greedy} and~\ref{tab:enhanced}. The curves show diminishing returns overall: substantial improvements are obtained in the first 10--100 reconstruction iterations, although CVRP1000 continues to improve markedly up to 1,000 iterations. RRC1000 is the high-compute end of this anytime trajectory rather than a required operating point.

Runtime environment: Python 3.12.7; PyTorch 2.11.0+cu130; CUDA 13.0; CPU x86\_64 (44 PyTorch threads); GPU NVIDIA RTX PRO 6000 Blackwell Max-Q Workstation Edition.

\section{Additional results}
\label{app:results}

\begin{table}[h]
\caption{Target-size calibration ablation for S3 (greedy, mean per-instance gap to LKH3, \%). Fixed $\alpha$ uses the CVRP100-selected value $\alpha=100$ for all sizes.}
\label{tab:alpha}
\centering
\begin{tabular}{@{}lrrrrr@{}}
\toprule
Inference & CVRP100 & CVRP200 & CVRP500 & CVRP1000 & Mean \\
\midrule
size-calibrated $\alpha$ & 1.019 & 0.828 & 1.407 & 7.544 & 2.700 \\
fixed $\alpha$ & 1.019 & 0.828 & 1.358 & 7.694 & 2.725 \\
\bottomrule
\end{tabular}
\end{table}

\begin{table}[h]
\caption{Full enhanced-inference results on standard CVRP, including all reported RRC budgets of LEHD and MnLP (mean per-instance gap to LKH3, \%).}
\label{tab:enhanced_full}
\centering
\setlength{\tabcolsep}{2.8pt}
\begin{tabular}{@{}llrrrrrl@{}}
\toprule
Method & Inference & CVRP100 & CVRP200 & CVRP500 & CVRP1000 & Mean & Source \\
\midrule
MDAM & bs50 & 2.211 & 4.304 & 10.498 & 27.814 & 11.207 & ReLD Table 3 \\
POMO & aug$\times$8 & 1.004 & 3.403 & 11.135 & 110.632 & 31.544 & ReLD Table 3 \\
ELG & aug$\times$8 & 1.207 & 2.553 & 5.472 & 10.760 & 4.998 & ReLD Table 3 \\
ReLD & aug$\times$8 & 0.960 & 1.654 & 2.975 & 6.757 & 3.087 & ReLD Table 3 \\
BQ & bs16 & 0.611 & 1.141 & 2.991 & 7.784 & 3.132 & LEHD Table 1 \\
BQ & bs16, later rerun & 1.020 & 0.940 & 1.010 & 2.880 & 1.463 & DRHG Table 3 \\
LEHD & RRC50 & 0.535 & 0.515 & 0.930 & 2.814 & 1.199 & LEHD Table 1 \\
LEHD & RRC100 & 0.272 & 0.217 & 0.546 & 2.370 & 0.851 & LEHD Table 1 \\
LEHD & RRC300 & 0.029 & $-$0.146 & 0.045 & 1.582 & 0.378 & LEHD Table 1 \\
LEHD & RRC500 & $-$0.044 & $-$0.246 & $-$0.107 & 1.270 & 0.218 & LEHD Table 1 \\
LEHD & RRC1000 & $-$0.112 & $-$0.383 & $-$0.347 & 0.921 & 0.020 & LEHD Table 1 \\
MnLP & RRC100 & 0.471 & 0.363 & 0.627 & 2.859 & 1.080 & MnLP Table 1 \\
MnLP & RRC500 & 0.094 & $-$0.136 & 0.007 & 1.624 & 0.397 & MnLP Table 1 \\
MnLP & RRC1000 & 0.012 & $-$0.255 & $-$0.257 & 1.148 & 0.162 & MnLP Table 1 \\
DRHG & $T{=}1000$ & $-$0.020 & $-$0.160 & 1.310 & 7.660 & 2.198 & DRHG Table 3 \\
\midrule
S0 & RRC1000 & 0.597 & 0.400 & 0.658 & 3.125 & 1.195 & This work \\
S1 & RRC1000 & 0.018 & 0.144 & 0.620 & 1.957 & 0.685 & This work \\
S2 & RRC1000 & $-$0.199 & $-$0.605 & $-$1.102 & $-$0.014 & $-$0.480 & This work \\
S3 & RRC1000 & \textbf{$-$0.216} & \textbf{$-$0.722} & \textbf{$-$1.179} & \textbf{$-$0.272} & \textbf{$-$0.597} & This work \\
\bottomrule
\end{tabular}
\end{table}

\end{document}